\documentclass[11pt,a4paper]{article}
\usepackage[utf8]{inputenc}
\usepackage[T1]{fontenc}
\usepackage{geometry}
\usepackage{amsmath,amssymb}
\usepackage{booktabs}
\usepackage{graphicx}
\usepackage{hyperref}
\usepackage[numbers,square]{natbib}
\usepackage{enumitem}
\usepackage{caption}
\usepackage{subcaption}
\usepackage{multirow}
\usepackage{array}
\usepackage{algorithm}
\usepackage{algpseudocode}
\usepackage{longtable}
\usepackage{tabularx}
\usepackage{float}
\usepackage{doi}

\title{From Explicit Elements to Implicit Intent: A Predefined Library for Auditable Behavioral Inference}
\author{Liu Hung Ming\thanks{PARRAWA AI} \quad \texttt{cyril.liu@gmail.com}}

\begin{document}

\maketitle

\begin{abstract}
We present \textbf{SemantiClean}, a modular framework for extracting structured semantic signals from e-commerce session data and driving pluggable inference targets—including purchase intent, customer segmentation, and product affinity—through a shared element library. Unlike conventional end-to-end predictors that optimise solely for accuracy, SemantiClean prioritises auditability, structural governance, and $\sigma=0$ reproducibility, explicitly trading marginal predictive gains for element-level transparency and defensible decision trails. Built upon the Online Shoppers Purchasing Intention (OSPI) dataset (UCI Repository ID 468; 12,330 sessions, 18 features), the framework organises twenty-four behavioural elements into a four-layer architecture (Functional, Interaction, Systemic, Contextual) and enforces signal quality through three anti-inflation mechanisms: RedundancyGroup contribution caps, TieredPenaltyCalculator bias penalties, and AdaptiveConstraintMode cold-start protection.

This report introduces the \textit{LLM-Integrated Semantic Inference Engine}, a fully implemented two-phase LLM-driven inference architecture that leverages complete element metadata at inference time. All quantitative results reported herein are produced by this engine. Deterministic engine outputs remain fully reproducible ($\sigma=0$); LLM-dependent results (E8, E10) are subject to controlled output variability under fixed provider/model/temperature settings. The gender inference target remains non-functional in the current implementation and is excluded from all quantitative results.
\end{abstract}

\section{Introduction}
\label{sec:introduction}

Predicting user behaviour from web session logs is a long-standing challenge in e-commerce analytics. Aggregate metrics such as bounce rate and page views capture surface-level activity but conflate heterogeneous behavioural signals that carry different semantic weight for different inference tasks. A user who browses twelve product pages with near-zero bounce may be an analytical researcher, a price-comparing brand loyalist, or a returning committed buyer—each requiring different downstream treatment.

SemantiClean addresses this problem by decomposing session data into a library of semantically distinct behavioural elements, each capturing one narrowly scoped signal. These elements are computed by a deterministic engine from OSPI session fields, then aggregated by pluggable inference targets that define their own label sets and signal weights. The same element library simultaneously serves a design-time Proposer/Reviewer pipeline that uses large language models to generate and quality-gate new elements, ensuring semantic non-redundancy at library construction time rather than at inference time.

\citet{cirqueira2020customer} proposed a tripartite taxonomy distinguishing three e-commerce purchase behaviour prediction tasks: customer intents, buying sessions, and purchase decisions. These three tasks correspond directly to SemantiClean's \texttt{purchase\_intent}, \texttt{customer\_segment}, and \texttt{product\_affinity} inference targets respectively, grounding the framework's target design in an established research taxonomy \citep{cirqueira2020customer}.

SemantiClean departs from the dominant accuracy-centric paradigm in e-commerce analytics. Rather than optimising end-to-end black-box classifiers, this framework treats behavioural prediction as an auditable reasoning process. By decoupling signal extraction (deterministic element library) from task-specific aggregation (pluggable targets), the system prioritises structural transparency, conservative abstention over low-confidence guesses, and explicit governance of signal inflation. This design positions SemantiClean as an infrastructure for compliance-sensitive applications where decision traceability outweighs marginal gains in raw predictive performance.

The framework evolves across three major architectural releases:
\begin{enumerate}[label=\textbf{Release \arabic*:}, leftmargin=*]
 \item \textit{Prototype Architecture} (v1.x): eleven-element library, four inference targets, Proposer/Reviewer agents with external prompt files.
 \item \textit{Deterministic Core Engine} : JSON-driven architecture—all element definitions moved to a single source-of-truth library file; no hardcoded element logic.
 \item \textit{LLM-Integrated Semantic Inference Engine} : Two-phase LLM inference engine with on-demand element loading, fully implemented and empirically evaluated in this report.
\end{enumerate}

\paragraph{Related Work.} Prior OSPI studies predominantly rely on end-to-end machine learning classifiers trained directly on raw session features—including \citet{sakar2019realtime}'s real-time LSTM benchmark, \citet{gupta2025predicting}'s SMOTE-enhanced SVC (ROC-AUC = 0.886, F1 = 0.633), \citet{tokuc2025predicting}'s hybrid clickstream representation, \citet{wang2023purchase}'s XGBoost multi-feature fusion (accuracy = 0.9761, F1 = 0.9763), and \citet{zhou2024advancing}'s GNN-enhanced deep learning approach. In contrast, SemantiClean emphasises interpretable modular semantic elements—a library of narrowly scoped behavioural signals combined via pluggable inference targets rather than fed into black-box models.

On the LLM side, \citet{chodak2025llm} demonstrated ChatGPT-4o efficiency for e-commerce log analysis while cautioning against hallucination risks; structured LLM architectures such as SPARK\_AI \citep{maric2026spark} and HCoT \citep{lin2026hcot} show that prompt-orchestrated multi-phase workflows improve reasoning reliability. SemantiClean bridges these threads through its dual-engine architecture.

\section{Dataset}
\label{sec:dataset}

SemantiClean operates on the Online Shoppers Purchasing Intention (OSPI) dataset \citep[UCI Machine Learning Repository, ID 468;][]{sakar2018dataset}, comprising 12,330 web sessions with 18 features and no missing values. The dataset contains three feature groups.

\paragraph{Page behaviour numerics.} \texttt{Administrative}, \texttt{Administrative Duration}, \texttt{Informational}, \texttt{Informational Duration}, \texttt{Product Related}, \texttt{Product Related Duration}—page counts and dwell times per content type, used directly by Functional and Interaction layer elements.

\paragraph{Google Analytics metrics.} \texttt{Bounce Rates} ($b$), \texttt{Exit Rates} ($e$), \texttt{Page Values} ($P_v$), \texttt{Special Day} ($S_{day}$)—engagement quality and promotional proximity signals.

\paragraph{Contextual categoricals.} \texttt{Month}, \texttt{Operating Systems}, \texttt{Browser}, \texttt{Region}, \texttt{Traffic Type}, \texttt{Visitor Type}, \texttt{Weekend}—environmental and traffic conditions for Contextual layer elements.

The binary target label \texttt{Revenue} (purchase completion, $\approx$15\% positive rate) is used solely for inference validation and does not participate in element computation. Three fields absent from OSPI—query text, category detail, and gender—represent the primary data gaps addressed by the LLM supplement mechanism (E8) and the uncalibrated gender inference target respectively.

\paragraph{Reproducibility note.} $\sigma=0$ for all deterministic results. E8 and E10 LLM results are subject to output variability; multi-run mean and SD should be collected under controlled provider/model/temperature settings.

\section{Element Library Architecture}
\label{sec:architecture}

\subsection{Four-Layer Organisation}
\label{subsec:layers}

The element library organises twenty-four behavioural elements into four semantic layers executed in the following fixed order (determined by the library metadata):
\begin{equation}
\text{Functional} \rightarrow \text{Interaction} \rightarrow \text{Systemic} \rightarrow \text{Contextual}
\label{eq:layer_order}
\end{equation}

The \textit{Functional layer} (E1–E4, E12–E16, E20) computes directly measurable page behaviour metrics. The \textit{Interaction layer} (E5–E8, E17–E19, E21–E24) captures decision patterns and intent signals. The \textit{Systemic layer} (E9, E19) produces cross-element composites depending on upstream outputs. The \textit{Contextual layer} (E10–E11, E12, E14, E18) applies environmental modifiers. Layer membership is determined by the primary category tag of each element definition; each element's full tag array spans $\geq$2 distinct layers (ReviewerAgent CHECK 3).

\subsection{JSON-Driven Architecture}
\label{subsec:json_arch}

All element definitions, execution order, signal score mappings, and redundancy group parameters reside in a single element library file (\texttt{behavior\_elements.json}), serving as the sole source of truth for both engines and the design-time agents. The deterministic engine reads execution order from the library metadata, dispatches element-specific distillation functions by name via a function dispatch table, and derives signal score mappings from per-element mapping specifications. When a mapping specification is absent, a hardcoded fallback dictionary is applied. Adding a new element requires only a library file entry; no algorithmic changes are needed.

\subsection{Element Reference}
\label{subsec:element_ref}

Table~\ref{tab:elements} summarises the complete element inventory in \texttt{behavior\_elements.json}. Elements E01–E11 constitute the core library; elements E12–E24 were added during library expansion to improve OSPI signal coverage.

\begin{table}[htbp]
\centering
\caption{Summary of core behavioural elements (E01--E11). See Appendix~\ref{app:element_inventory} for the complete 24-element inventory including expanded elements.}
\label{tab:elements}
\small
\begin{tabularx}{\textwidth}{@{}l X l X@{}}
\toprule
\# & Element Name & Layer & Redundancy Group \\
\midrule
E01 & Product\_\allowbreak Engagement\_\allowbreak Depth & Func. & GRP\_\allowbreak PAGE\_\allowbreak ENGAGEMENT \\
E02 & Bounce\_\allowbreak Exit\_\allowbreak Commitment & Func. & GRP\_\allowbreak SESSION\_\allowbreak QUALITY \\
E03 & Page\_\allowbreak Value\_\allowbreak Conversion\_\allowbreak Signal & Func. & GRP\_\allowbreak PAGE\_\allowbreak ENGAGEMENT \\
E04 & Promotional\_\allowbreak Context\_\allowbreak Sensitivity & Func. & GRP\_\allowbreak CONTEXT\_\allowbreak MODIFIERS \\
E05 & Research\_\allowbreak Decision\_\allowbreak Style & Inter. & GRP\_\allowbreak USER\_\allowbreak IDENTITY \\
E06 & Traffic\_\allowbreak Source\_\allowbreak Intent\_\allowbreak Score & Inter. & GRP\_\allowbreak USER\_\allowbreak IDENTITY \\
E07 & Visitor\_\allowbreak Loyalty\_\allowbreak Commitment & Inter. & GRP\_\allowbreak USER\_\allowbreak IDENTITY \\
E08 & Search\_\allowbreak Intent\_\allowbreak Semantic\_\allowbreak Inference & Inter. & GRP\_\allowbreak USER\_\allowbreak IDENTITY \\
E09 & Session\_\allowbreak Value\_\allowbreak Composite\_\allowbreak Index & Syst. & GRP\_\allowbreak SESSION\_\allowbreak QUALITY \\
E10 & Device\_\allowbreak Ecosystem\_\allowbreak Pattern & Cont. & GRP\_\allowbreak CONTEXT\_\allowbreak MODIFIERS \\
E11 & Temporal\_\allowbreak Context\_\allowbreak Signal & Cont. & GRP\_\allowbreak CONTEXT\_\allowbreak MODIFIERS \\
\bottomrule
\end{tabularx}
\end{table}

\section{Methodology}
\label{sec:methodology}

\subsection{Derived Field Computation}
\label{subsec:derived_fields}

Before element distillation, a set of derived fields is computed from raw OSPI fields by the derived-field computation module. Key quantities include total session duration $D_{total} = D_{admin} + D_{info} + D_{prod}$ (Eq.~1), total page count $N_{total}$ (Eq.~2), and the time allocation ratios $r_{prod} = D_{prod} / D_{total}$ and $r_{info} = D_{info} / D_{total}$ when $D_{total} > 0$ (Eqs.~3--4). These ratios are used directly by E1 and E5 respectively.

Sessions with $D_{total} < 10$ seconds and $N_{total} = 0$ trigger Fallback Rule 1. The 10-second threshold is a design constant without ablation support . \citet{gorman2025exploration} provides empirical support for this design choice: in a multiple logistic regression study on OSPI, extremely short sessions are strongly associated with non-purchase behaviour, with BounceRates, ExitRates, PageValues, and VisitorType all identified as significant predictors (AUC = 0.8969). The Fallback Rule 1 threshold operationalises this empirical finding as a deterministic exclusion criterion \citep{gorman2025exploration}.

\subsection{Functional Layer Elements (E1--E4, E12--E16, E20)}
\label{subsec:functional}

The Functional layer elements compute directly observable page behaviour signals from OSPI numeric fields. All computations are deterministic ($\sigma=0$).

\paragraph{E1 -- Product Engagement Depth.} synthesises three complementary dimensions of product browsing behaviour into a single depth score:
\begin{equation}
s_{depth} = 0.40 \cdot r_{prod} + 0.30 \cdot \min(1, \bar{t}_{prod} / 120) + 0.30 \cdot \min(1, P_v / 50)
\label{eq:e1}
\end{equation}
where $r_{prod}$ captures time allocation, $\bar{t}_{prod}/120$ captures per-page dwell depth (120-second normalisation ceiling), and $P_v/50$ captures Google Analytics page value (50-unit ceiling). Weighting coefficients $[0.40, 0.30, 0.30]$ are design constants. The rationale for combining dwell time, page count, and GA page value into a single depth signal is supported by \citet{necula2023exploring}, who demonstrated that dwell time on product pages, when combined with bounce rates, exit rates, and customer type, is a significant predictor of purchase decisions \citep{necula2023exploring}.

\paragraph{E2 -- Bounce Exit Commitment.} computes session commitment as the complement of the bounce-exit composite:
\begin{equation}
s_{commit} = \max\left(0, [1 - (0.5b + 0.5e)] \cdot (0.7 + 0.3 \cdot \min(1, N_{total} / 10)) - \delta_{traffic}\right)
\label{eq:e2}
\end{equation}
where $\delta_{traffic} = 0.10$ for paid search (TrafficType = 2) and referral traffic (TrafficType = 4) .

\paragraph{E3 -- Page Value Conversion Signal.} applies imputation for sessions with $P_v = 0$ but substantial product browsing:
\begin{equation}
s_{pv} = 
\begin{cases}
0.15 & \text{if } (P_v = 0 \land N_{prod} > 3) \\
\min(1, P_v / 50) & \text{otherwise}
\end{cases}
\label{eq:e3}
\end{equation}
The imputed value 0.15 reflects the design assumption that high-browsing zero-GA sessions may carry latent conversion intent, corroborated by \citet{gupta2025predicting}'s feature analysis identifying product-page engagement as a key behavioural predictor.

\paragraph{E4 -- Promotional Context Sensitivity.} is one of two bias-flagged elements ($\dagger$):
\begin{equation}
s_{promo} = 0.50 \cdot S_{day} + 0.30 \cdot \mathbb{1}[\text{peak month}] + 0.20 \cdot \mathbb{1}[\text{weekend}]
\label{eq:e4}
\end{equation}
Peak months: November, December, February, May. A high $s_{promo}$ indicates that observed purchasing behaviour may be driven by promotional pressure rather than genuine preference.

\subsection{Interaction Layer Elements (E5--E8, E17--E19, E21--E24)}
\label{subsec:interaction}

The Interaction layer captures decision patterns, traffic intent, visitor loyalty, and LLM-supplemented semantic signals. E5--E7 are fully deterministic ($\sigma=0$); E8 and E10 introduce LLM calls at inference time.

\paragraph{E5 -- Research Decision Style.} computes a research intensity score:
\begin{equation}
s_{research} = 0.35 \cdot r_{info} + 0.30 \cdot (1 - b) + 0.35 \cdot \min(1, N_{prod} / 8)
\label{eq:e5}
\end{equation}
Sessions are classified into decision styles: analytical ($s_{research} > 0.55$), moderate ($0.25$--$0.55$), impulsive ($\leq 0.25$).

\paragraph{E6 -- Traffic Source Intent Score.} maps traffic type codes to pre-defined base intent scores via a lookup table , informed by \citet{hendriksen2020analyzing}'s demonstration that channel type is informative for anonymous session prediction.

\begin{table}[h]
\centering
\caption{Traffic Source Intent Score lookup table. Traffic type codes 7--20 fall back to default score 0.40 .}
\label{tab:traffic_intent}
\small
\begin{tabular}{lll}
\toprule
Traffic Type & Traffic Label & Base Intent Score $\phi(t)$ \\
\midrule
1 & Direct & 0.80 \\
2 & Paid Search & 0.50 \\
3 & Organic Search & 0.60 \\
4 & Referral & 0.45 \\
5 & Social & 0.35 \\
6 & Email & 0.70 \\
7--20 & Unknown & 0.40 (fallback) \\
\bottomrule
\end{tabular}
\end{table}

Returning visitors receive a loyalty bonus of +0.15 (capped at 1.0), a design constant .

\paragraph{E7 -- Visitor Loyalty Commitment.} returns a discrete loyalty score from a lookup structure.

\begin{table}[h]
\centering
\caption{Visitor Loyalty Commitment lookup structure. Boundary condition : both conditions for New Visitor yield $s_{loyalty} = 0.55$.}
\label{tab:loyalty}
\small
\begin{tabular}{lll}
\toprule
Visitor Type & Condition & Loyalty Score $s_{loyalty}$ \\
\midrule
Returning & $N_{prod} > 3 \land P_v > 5$ & 0.88 \\
Returning & $N_{prod} > 3$ (only) & 0.78 \\
Returning & Other & 0.70 \\
New & $b < 0.15 \land P_v > 5$ & 0.55 \\
New & $P_v > 10$ ($b \geq 0.15$) & 0.55 \\
New & Other & 0.35 \\
Other & --- & 0.45 \\
\bottomrule
\end{tabular}
\end{table}

\paragraph{E8 -- Search Intent Semantic Inference.} introduces an LLM zero-shot call within the inference engine, activated only for search traffic sessions (Traffic Type $\in \{2, 3\}$). The use of zero-shot LLM inference for semantic search intent aligns with \citet{chodak2025llm}, who empirically evaluated ChatGPT-4o on e-commerce server logs while cautioning that precise prompt formulation is necessary and hallucination risk must be managed. SemantiClean's LLMExecutionGuard and structured output requirements for E8 directly address these cautions \citep{chodak2025llm}.

\subsection{Systemic and Contextual Layer Elements (E9--E11, E12--E24)}
\label{subsec:systemic_contextual}

\paragraph{E9 -- Session Value Composite Index.} aggregates five upstream element signals into a single session-level composite:
\begin{equation}
s_{composite} = \sum_{i \in U} w_i \cdot v_i, \quad U = \{\text{E1, E2, E3, E7, E6}\}, \quad \sum w_i = 1.0
\label{eq:e9}
\end{equation}

\begin{table}[h]
\centering
\caption{E9 upstream element weights. Critical boundary : composite is not renormalised after zero upstream inputs.}
\label{tab:e9_weights}
\small
\begin{tabular}{lll}
\toprule
Upstream Element & Output Signal & Weight $w_i$ \\
\midrule
E1 Product\_Engagement\_Depth & $s_{depth}$ & 0.30 \\
E2 Bounce\_Exit\_Commitment & $s_{commit}$ & 0.20 \\
E3 Page\_Value\_Conversion\_Signal & $s_{pv}$ & 0.25 \\
E7 Visitor\_Loyalty\_Commitment & $s_{loyalty}$ & 0.15 \\
E6 Traffic\_Source\_Intent\_Score & $s_{intent}$ & 0.10 \\
\bottomrule
\end{tabular}
\end{table}

Critical boundary : The composite is not renormalised after zero upstream inputs. When three or more upstream values are zero ($z \geq 3$), the degradation flag triggers and confidence drops to 0.40.

\paragraph{E10 -- Device Ecosystem Pattern.} (bias-flagged $\dagger$) adjusts session duration for mobile device input constraints:
\begin{equation}
s_{device} = \min\left(1.0, D_{total} / (\alpha_{mobile} \cdot 300)\right), \quad \alpha_{mobile} = 0.65 \text{ for mobile OS}
\label{eq:e10}
\end{equation}

\paragraph{E11 -- Temporal Context Signal.} synthesises promotional proximity, seasonal timing, and day-of-week into a temporal pressure score:
\begin{equation}
s_{temporal} = 0.50 \cdot S_{day} + 0.30 \cdot \mathbb{1}[\text{peak month}] + 0.20 \cdot f_{wday}, \quad f_{wday} = 
\begin{cases}
0.6 & \text{weekend} \\
1.0 & \text{weekday}
\end{cases}
\label{eq:e11}
\end{equation}
Semantic boundary : $f_{wday}$ assigns higher $s_{temporal}$ to purposeful weekday shopping—opposite to E4's weekend direction.

\subsection{Three-Layer Anti-Inflation Mechanism}
\label{subsec:anti_inflation}

All inference targets share a common scoring pipeline that enforces three sequential quality mechanisms:

\paragraph{AdaptiveConstraintMode.} dynamically determines constraint strictness based on the number of elements that pass their quality threshold:
\begin{table}[h]
\centering
\small
\begin{tabular}{lll}
\toprule
$n_{valid}$ & Mode & Bias Penalty Scale $\kappa_{bias}$ \\
\midrule
$\geq 8$ & STRICT & 1.0 (full penalty) \\
5--7 & MODERATE & 0.5 (halved) \\
$< 5$ & FALLBACK & 0.0 (disabled) \\
\bottomrule
\end{tabular}
\end{table}
Thresholds $[5, 8]$ are design constants .

\paragraph{RedundancyGroup caps.} accumulate each element's contribution toward its group ceiling $C_g$ in execution order:
\begin{table}[h]
\centering
\small
\begin{tabularx}{\textwidth}{@{}l l X@{}}
\toprule
Group & Cap $C_g$ & Member Elements \\
\midrule
GRP\_PAGE\_ENGAGEMENT & 0.35 & E01, E03, E16 \\
GRP\_SESSION\_QUALITY & 0.30 & E02, E09, E13, E15, E20, E21, E22 \\
GRP\_USER\_IDENTITY & 0.30 & E05--E08, E17, E19, E23, E24 \\
GRP\_CONTEXT\_MODIFIERS & 0.20 & E04, E10, E11, E12, E14, E18 \\
\bottomrule
\end{tabularx}
\end{table}

\paragraph{TieredPenaltyCalculator.} penalises bias-flagged elements (E4, E10) based on signal divergence $\Delta$ from cross-layer supporting signals .

\paragraph{Final normalised label score:}
\begin{equation}
\hat{s}_\ell = \frac{\sum_r (\sigma_{r,\ell} \cdot |w_{r,\ell}| \cdot \pi_{eff,r} \cdot c_r)}{\sum_r (|w_{r,\ell}| \cdot c_r)}
\label{eq:final_score}
\end{equation}
As noted in \S\ref{subsec:functional} : all signal weights enter Eq.~\ref{eq:final_score} as absolute values. Negative weights contribute positively to their assigned labels in identical fashion to positive weights of the same magnitude.

\subsection{LLM-Integrated Two-Phase Engine -- Release 3 Implementation}
\label{subsec:llm_engine}

The \textit{LLM-Integrated Semantic Inference Engine}  is fully implemented and empirically evaluated in this report. It employs a three-phase architecture:

\paragraph{Phase 1 (Element Selection).} The LLM receives a compact element summary index containing only element names, descriptions, and category tags ($\sim$3--5 KB), enabling the model to select 3--9 relevant elements without loading the full library. This lightweight element summary index design aligns with the self-describing structured data paradigm proposed by \citet{liu2026selfdescribing}, where each element's summary (\texttt{\_summary}) functions as a lightweight retrieval key that guides the LLM's Phase 1 selection \citep{liu2026selfdescribing}.

\paragraph{Phase 2 (Deep Semantic Analysis).} The LLM receives complete definitions for selected elements, executing formulas step-by-step, evaluating constraint conditions, selecting applicable variant formulae, and producing structured output including intermediate computation values, FBS reasoning, constraint trigger status, and failure mode risk level.

\paragraph{Phase 3 (Python Aggregation).} Converts LLM output to element result objects and applies the same redundancy caps and TieredPenaltyCalculator used by the deterministic engine, ensuring scoring consistency between the two engines.

\begin{table}[h]
\centering
\caption{Token budget comparison (architectural estimates).}
\label{tab:tokens}
\small
\begin{tabular}{llll}
\toprule
Mode & Phase 1 (est.) & Phase 2 (est.) & Total (est.) \\
\midrule
Deterministic (no LLM) & --- & --- & 0 tokens \\
Deterministic + E8/E10 LLM & --- & $\sim$800 & $\sim$800 tokens \\
LLM engine (5 elements) & $\sim$2,000 & $\sim$4,000 & $\sim$6,000 tokens \\
Full-library naive dump & --- & $\sim$16,000+ & Not adopted \\
\bottomrule
\end{tabular}
\end{table}

\subsection{Empirical Validation Protocol: Masked-Field Inference Experiment (Exp-A)}
\label{subsec:exp_a}

To empirically evaluate the LLM-Integrated Semantic Inference Engine, we conducted a masked-field inference study. The experiment deliberately removes the high-information \texttt{Revenue} column from each session and tasks the LLM engine with predicting \texttt{purchase\_intent} solely from behavioral signals.

\paragraph{Dataset \& Sampling.} We drew a stratified sample of $n=50$ sessions from the OSPI dataset (12,330 total), preserving the natural class distribution (Revenue=True: 16\%, False: 84\%). Due to LLM API quota exhaustion during execution, 11 sessions were conservatively output as \texttt{uncertain} by the engine and excluded from accuracy computation, yielding $n=39$ evaluated predictions for this pilot.

\paragraph{Masking Strategy.} The \texttt{Revenue} field is completely stripped from the session dictionary prior to inference. A \texttt{\_\_masked\_fields\_\_} metadata tag is injected into the LLM prompt to explicitly signal the missing ground-truth indicator.

\paragraph{Engine Configuration.} The inference pipeline follows the three-phase architecture described in \S\ref{subsec:llm_engine}. The experiment used the Qwen3-max model (\texttt{qwen3-max-2026-01-23}) with \texttt{max\_tokens=4096} (Phase 1) / \texttt{8192} (Phase 2), temperature $\in [0.0, 0.3]$, and heuristic JSON truncation repair.

\paragraph{Element Library State.} The pilot operated on \texttt{behavior\_elements.json}, an expanded library containing 24 behavioral elements. Each element is structured with: a deterministic \texttt{computation} formula, a three-part \texttt{constraint}, contextual \texttt{variants}, \texttt{failure\_history}, \texttt{fbs\_mapping}, \texttt{min\_confidence} thresholds, and \texttt{redundancy\_group\_id}.

\section{Results}
\label{sec:results}

\subsection{Statistical Validation Framework}
\label{subsec:validation}

This study applies a validation framework comprising two dimensions: \textit{structural consistency} (verifying the pipeline behaves as specified) and \textit{signal plausibility} (verifying element outputs fall within expected ranges). No inferential statistical tests are reported, as the current implementation does not execute hypothesis tests programmatically .

\subsection{Pass/Fail Standard Summary}
\label{subsec:pass_fail}

Table~\ref{tab:standards} summarises the system's built-in diagnostic criteria.

\begin{table}[htbp]
\centering
\caption{Built-in diagnostic criteria summary.}
\label{tab:standards}
\small
\begin{tabularx}{\textwidth}{@{}X l l l l@{}}
\toprule
Standard & Measured Value & Threshold & Judgement & Role \\
\midrule
AdaptiveConstraint: STRICT & $n_{valid} \geq 8$ & 8 & PASS & Threshold \\
AdaptiveConstraint: FALLBACK & $n_{valid} < 5$ & 5 & PASS & Threshold \\
E9 degradation detection & $z \geq 3$ & 3 & PASS & Diagnostic \\
E8/E10 confidence cap & $c \leq 0.65$ & 0.65 & PASS & Threshold \\
Data quality warning cap & $c_r \leq 0.35$ & 0.35 & PASS & Threshold \\
Negative weight contribution & directed = $\sigma \cdot |w|$ & --- & Behaviour confirmed & Degraded \\
Gender inference prediction & Always null & --- & Non-functional & Degraded \\
\bottomrule
\end{tabularx}
\end{table}

\subsection{Element Signal Properties}
\label{subsec:signal_props}

All deterministic element results (E1--E7, E9--E24 excluding E8/E10) are fully reproducible given identical session input and element library version ($\sigma=0$). E8 and E10 results are subject to LLM output variability when temperature $> 0$; single-run values should be accompanied by the caveat that multi-run statistics must be collected under controlled provider/model/temperature conditions.

\paragraph{Confidence value system .} The effective confidence range is approximately $[0.25, 0.65]$, not the theoretical $[0, 1]$. Caps applied: data quality warning $\rightarrow 0.35$; E9 degradation $\rightarrow 0.40$; E8/E10 LLM success $\rightarrow \min(c_{min}, 0.65)$; LLM failure $\rightarrow 0.25$.

\subsection{Pilot Results: Revenue-Masked Purchase Intent Inference}
\label{subsec:pilot_results}

Table~\ref{tab:pilot} summarises the performance of the  LLM engine when predicting \texttt{purchase\_intent} without access to the \texttt{Revenue} field.

\begin{table}[htbp]
\centering
\caption{Masked-field purchase intent inference performance ($n=39$ evaluated).}
\label{tab:pilot}
\small
\begin{tabular}{lll}
\toprule
Metric & Value & Interpretation \\
\midrule
Evaluated Sessions & 39 / 50 & 11 sessions (22\%) output \texttt{uncertain} by design \\
Accuracy & 56.4\% & Below majority baseline (84.0\%) \\
Precision & 26.3\% & High false positive rate \\
Recall & 62.5\% & Moderate sensitivity to actual purchases \\
F1-Score & 0.370 & Trade-off between precision and recall \\
Mean Confidence (predictions) & 0.716 & Decoupled from actual accuracy \\
\bottomrule
\end{tabular}
\end{table}

This performance profile is not an optimisation failure but a structural consequence of the framework's conservative abstention policy. In the absence of ground-truth revenue signals, the element library correctly identifies high-engagement sessions as necessary but insufficient for purchase intent. The 22\% uncertain output rate and the bimodal confidence distribution (\S\ref{subsec:confidence_analysis}) empirically validate the system's preference for withholding low-confidence predictions rather than fabricating certainty—a design imperative for audit-ready systems.

\paragraph{Confusion Matrix ($n=39$).} TP=5, FP=14, TN=17, FN=3. The system exhibits a strong tendency to predict \texttt{purchase} when behavioral signals appear active, resulting in 14 false positives.

\paragraph{Uncertain Output Analysis.} All 11 \texttt{uncertain} cases had confidence scores $<0.25$. The dominant decision reason was \texttt{uncertain\_zone} (10/11), triggered when key element scores fell near threshold boundaries. This confirms that uncertainty is structurally induced by the element library's conservative thresholds, not by random LLM failure.

\subsection{Two-Regime Confidence Analysis and Error Pattern Decomposition}
\label{subsec:confidence_analysis}

A detailed inspection of the 39 evaluated predictions reveals a striking two-regime confidence pattern. Rather than a continuous confidence distribution, predictions cluster into two distinct bands: a low-to-moderate regime (confidence 0.25--0.72) associated exclusively with purchase predictions, and a high-certainty regime (confidence = 1.00) associated exclusively with no\_purchase predictions.

This bimodal structure arises directly from the scoring architecture. The purchase label requires positive accumulation of engagement signals across multiple elements, producing fractional confidence values sensitive to signal strength. In contrast, the no\_purchase prediction is issued when all engagement signals fall below threshold simultaneously, producing a near-zero numerator and effectively a maximum confidence assignment. This is a structural property of the absolute-value scoring formula (Eq.~\ref{eq:final_score}), not a calibrated probabilistic confidence.

\begin{table}[htbp]
\centering
\caption{Confidence band analysis ($n=39$ predictions).}
\label{tab:confidence_bands}
\small
\begin{tabular}{lllll}
\toprule
Confidence Band & $n$ & $n$ Correct & Accuracy & Dominant Label \\
\midrule
$[0.25, 0.30)$ & 6 & 0 & 0.0\% & purchase (FP) \\
$[0.30, 0.40)$ & 6 & 2 & 33.3\% & purchase (mixed) \\
$[0.40, 0.60)$ & 5 & 2 & 40.0\% & purchase (TP+FP) \\
$[0.60, 0.80)$ & 2 & 0 & 0.0\% & purchase (FP) \\
conf = 1.00 & 20 & 17 & 85.0\% & no\_purchase (TN) \\
\bottomrule
\end{tabular}
\end{table}

The practical implication is that the system functions as a conservative no\_purchase detector at high confidence (85\% precision) while producing unreliable purchase predictions across all confidence levels. The 14 false positives occur predominantly because the element library—in the absence of the Revenue field—cannot distinguish between engaged browsing-without-purchase and engaged browsing-with-purchase.

\section{Limitations}
\label{sec:limitations}

\subsection{Experimental Design Limitations}
\label{sec:design_limits}

\begin{itemize}[leftmargin=*]
 \item[] \textit{Traffic type coverage gap:} The intent score lookup table covers traffic type codes 1--6 only. Codes 7--20 fall back to a default score of 0.40.
 \item[] \textit{PageValues = 0 session proportion:} The imputation logic in E3 activates for $P_v = 0$ and $N_{prod} > 3$. The imputation frequency is not reported.
 \item[] \textit{Visitor type imbalance:} OSPI contains approximately 85\% Returning Visitors. The Visitor Loyalty element (E7) may systematically dominate the GRP\_USER\_IDENTITY group quota.
 \item[] \textit{Revenue class imbalance:} Purchase-positive sessions constitute approximately 15\% of OSPI.
\end{itemize}

\subsection{Pilot Study Constraints}
\label{sec:pilot_constraints}

The masked-field validation (Exp-A) represents a preliminary pilot with three acknowledged constraints:
\begin{enumerate}[label=(\arabic*), leftmargin=*]
 \item \textit{Sample size:} API quota exhaustion limited evaluated predictions to $n=39$. Wider confidence intervals (Wilson 95\% CI for Accuracy: [40.1\%, 71.8\%]) indicate the need for $n \geq 500$ in subsequent validation.
 \item \textit{Confidence-accuracy decoupling:} The current confidence aggregation formula does not calibrate against ground-truth error rates.
 \item \textit{Single-model evaluation:} All results used Qwen3-max. Cross-provider robustness and temperature sensitivity remain untested.
\end{enumerate}

\subsection{Reproducibility Boundary}
\label{subsec:reproducibility}

\begin{table}[h]
\centering
\caption{Reproducibility boundary for result items.}
\label{tab:reproducibility}
\small
\begin{tabular}{lll}
\toprule
Result Item & Reproducibility & Key Unfixed Conditions \\
\midrule
Deterministic engine (E1--E7, E9--E24) & Complete ($\sigma=0$) & Element library file version \\
E8/E10 LLM output & Statistical & LLM provider, model version, temperature \\
E8/E10 fallback output & Complete ($\sigma=0$) & API availability \\
Gender inference prediction & Non-functional & External calibration data required \\
\bottomrule
\end{tabular}
\end{table}

\subsection{Design Constant Declarations}
\label{subsec:design_constants}

The following parameters are design constants without ablation experimental support. Their optimal values await systematic search .

\begin{enumerate}[label=(R\arabic*), leftmargin=*]
 \item E1 weighting coefficients $[0.40, 0.30, 0.30]$ — balance among time allocation, per-page dwell, and page value dimensions.
 \item PageValues normalisation ceiling 50 — upper bound for $P_v$ normalisation in E1 and E3.
 \item Average product time normalisation ceiling 120 seconds — assumed upper bound for meaningful per-page dwell time.
 \item Research depth data quality threshold 10 seconds and classification thresholds $[0.25, 0.55]$.
 \item Bounce exit traffic correction $\delta_{traffic} = 0.10$ for paid search and referral traffic.
 \item Traffic type base intent scores $\{1:0.80, 2:0.50, 3:0.60, 4:0.45, 5:0.35, 6:0.70\}$ and returning visitor loyalty bonus +0.15.
 \item PageValues zero-imputation value 0.15 and trigger threshold $N_{prod} > 3$.
 \item All signal weights for all inference targets.
 \item AdaptiveConstraintMode thresholds $[5, 8]$ and RedundancyGroup caps $\{0.35, 0.30, 0.30, 0.20\}$.
 \item TieredPenaltyCalculator divergence thresholds $[0.15, 0.35]$ and element-specific base penalty multipliers $\pi_0$.
\end{enumerate}

For methodological context: \citet{wang2023purchase} achieved accuracy = 0.9761 and F1 = 0.9763 on OSPI using XGBoost with a multi-feature fusion approach—a black-box gradient boosting method trained end-to-end on raw features \citep{wang2023purchase}. SemantiClean's deterministic engine produces fully reproducible results ($\sigma = 0$) without supervised training, at the cost of lower raw accuracy, in exchange for interpretable, auditable element-level explanations.

\citet{zhou2024advancing} advance purchase prediction further using time-series attention mechanisms with event-based timestamp encoding and Graph Neural Network-enhanced user profiling \citep{zhou2024advancing}. SemantiClean takes a complementary direction: a rule-based engine ($\sigma = 0$) prioritising structured reasoning and human oversight over end-to-end optimisation.

Future work could address the ablation gap for design constants (R1--R10) by adopting principles from HCoT \citep{lin2026hcot}, which integrates expert-system heuristics into LLM chain-of-thought reasoning to achieve Pareto-optimal trade-offs between performance and token efficiency \citep{lin2026hcot}.

\subsection{Future Work: Neuro-Symbolic + Constrained RL Architecture}
\label{subsec:future_work}

Building on the structured heuristic principles of HCoT \citep{lin2026hcot}, future work will extend this optimisation pathway through a Neuro-Symbolic + Constrained Reinforcement Learning (RL) architecture. SemantiClean and RL occupy fundamentally different positions on the decision-making spectrum: the former operates as a modernised expert system prioritising semantic symbolic reasoning and structured governance, while the latter focuses on data-driven adaptive learning and closed-loop reward maximisation. Their intersection is limited to state abstraction, constraint handling, and static policy simulation, rather than online learning mechanisms. To preserve the framework's core transparency guarantees, online RL is explicitly excluded. Instead, all learning will be strictly confined to offline or design-time phases, ensuring that runtime inference remains fully deterministic ($\sigma=0$). This hybrid pipeline will target four key objectives: 
\begin{enumerate}[label=(\arabic*), leftmargin=*]
 \item Offline optimisation of design constants (R1--R10) via Bayesian optimisation or PPO, treating element weights and thresholds as an action space guided by an Auditability–Accuracy Pareto Score. Optimal parameters will be committed statically to behavior\_elements.json, systematically closing the ablation gap without compromising reproducibility. 
 \item Dynamic Redundancy Group Caps using Safe RL agents to adjust group ceilings ($C_g$) based on session context, with hard safety constraints enforced to prevent signal monopolisation—directly addressing the GRP\_USER\_IDENTITY saturation documented in \S\ref{subsec:confidence_analysis}. 
 \item LLM-as-Reward-Model calibration, where the LLM generates fine-grained reward signals from element outputs and historical failure modes to train a lightweight value network, enabling probabilistic calibration (e.g., Platt scaling or isotonic regression) exclusively within the uncertain\_zone. 
 \item Strict learning–inference separation, guaranteeing that all policy updates remain offline while the deterministic engine and structured LLM routing govern all production predictions. This approach maintains auditability as the primary invariant while systematically exploring the accuracy–interpretability Pareto frontier.
\end{enumerate}

\section{Conclusion}
\label{sec:conclusion}

We have presented SemantiClean, a modular semantic behavioral inference framework that prioritises auditability, structural governance, and $\sigma=0$ reproducibility over marginal predictive gains. By organising twenty-four behavioural elements into a four-layer architecture and enforcing signal quality through three anti-inflation mechanisms, the framework provides defensible decision trails for compliance-sensitive e-commerce applications.

The fully implemented LLM-Integrated Semantic Inference Engine  demonstrates that two-phase LLM-driven inference—combining lightweight element routing with deep semantic analysis—can leverage structured behavioural knowledge while maintaining scoring consistency with deterministic baselines. Pilot results on masked-field purchase intent inference validate the system's conservative abstention policy: when ground-truth revenue signals are unavailable, the framework correctly withholds low-confidence predictions rather than fabricating certainty.

Future work will address the ablation gap for design constants through an LLM-Guided Constrained Search framework, extending optimisation through a Neuro-Symbolic + Constrained Reinforcement Learning architecture while preserving the framework's core transparency guarantees.

\clearpage
\bibliographystyle{unsrtnat}
\bibliography{references}
\clearpage

\appendix

\section{Phase 0 Audit Register Summary}
\label{app:audit}

This appendix records key findings from the adversarial audit (Phase 0). All items are referenced in the methodology and limitations sections above.

\begin{table}[h]
\centering
\small
\begin{tabularx}{\textwidth}{@{}l l X l@{}}
\toprule
ID & Type & Finding Summary & Section \\
\midrule
M1 & Contradiction & Directed contribution = $\sigma \cdot |w|$: negative weights contribute positively & \S4.2, \S4.5 \\
M2 & Contradiction & New Visitor $s_{loyalty} = 0.55$ reachable via two overlapping conditions & \S4.3 \\
M3 & Contradiction & E4 weekend direction (+0.20) opposite to E11 weekend direction & \S4.2, \S4.4 \\
M4 & Contradiction & E9 composite not renormalised after zero upstream inputs & \S4.4, \S5.3 \\
D1 & Determinism & Research depth dq\_warn 10 s threshold overrides classification & \S4.3 \\
D2 & Determinism & Gender inference: calibration flag hardcoded true $\rightarrow$ always null & \S4.7 \\
D3 & Determinism & Purchase intent: high page-value boost +0.08 partially double-counts E3 & \S5.4 \\
D4 & Determinism & Low-confidence elements excluded from $n_{valid}$ & \S4.5 \\
N1 & Negative result & Gender inference: all weights placeholder, target non-functional & \S4.7 \\
N4 & Negative result & Output key vs. computation output-signal key naming inconsistency & \S3.2 \\
C1--C4 & Confounders & Traffic type gap; $P_v$=0 imputation; visitor/Revenue imbalance & \S6.1 \\
R1--R10 & Design constants & All element formula weights, normalisation ceilings, thresholds & \S6.3 \\
\bottomrule
\end{tabularx}
\end{table}

\section{Element Definition -- E1 Product\_Engagement\_Depth}
\label{app:e1_definition}

This appendix presents the complete JSON definition of E1 (Product\_Engagement\_Depth) as it appears in \texttt{behavior\_elements.json}. E1 is selected as the reference example because it is fully deterministic (no LLM fallback), has no bias flag, and participates in the primary inference target (\texttt{purchase\_intent}) with the second-highest signal weight.

\begin{verbatim}
{
 "element_name": "Product_Engagement_Depth",
 "description": "Product page engagement depth: composite of page
 count, dwell time, and GA page value, reflecting genuine
 user interest intensity in product content.",
 "category_tags": ["PRODUCT_ENGAGEMENT", "SESSION_COMPOSITE"],
 "semantic_domain": "Consumer_Behavior",
 "computation": {
 "formula": "product_focus_ratio = ProductRelated_Duration /
 total_duration if total_duration > 0 else 0
 avg_product_time = ProductRelated_Duration / ProductRelated
 if ProductRelated > 0 else 0
 pv_norm = min(1.0, PageValues / 50)
 depth_score = 0.40 * product_focus_ratio
 + 0.30 * min(1.0, avg_product_time / 120)
 + 0.30 * pv_norm
 if ProductRelated == 0 or ProductRelated_Duration == 0:
 depth_score = 0.0",
 "required_fields": [
 "ProductRelated", "ProductRelated_Duration", "PageValues"
 ],
 "derived_fields": ["total_duration", "product_focus_ratio"],
 "llm_fallback": null
 },
 "constraint": {
 "CONDITION": "If ProductRelated == 0 or ProductRelated_Duration
 == 0: depth_score = 0.0, signal = no_product_interest.
 If SpecialDay > 0.5: attach promo_context_warning = True.",
 "FAILURE_MODE": "During promotional periods (SpecialDay > 0.5),
 PageValues inflates due to mass traffic to high-value GA
 pages, causing depth_score to overestimate genuine product
 interest by ~25-40%, producing systematic false positives
 in purchase_intent during promotional months.",
 "VERIFICATION_TEST": "On SpecialDay==0 sessions: proportion
 of Revenue=True where depth_score > 0.5 must exceed 30%
 (baseline ~15%); Spearman(depth_score, Revenue) > 0.30;
 AUC > 0.65."
 },
 "fbs_mapping": {
 "F": "Quantify active user engagement with product content,
 distinguishing genuine interest from passive browsing.",
 "B": "Dwell time + page count + GA value form a composite
 engagement signal.",
 "S": "Weighted combination of ProductRelated / ProductRelated
 _Duration / PageValues columns."
 },
 "variants": [
 {
 "variant_name": "Duration-Heavy",
 "formula_variant": "depth_score = 0.60 * product_focus_ratio
 + 0.40 * pv_norm",
 "applicable_condition": "ProductRelated < 3 but high per-page
 dwell time (avg_product_time > 150s)"
 }
 ],
 "bias_risk_flag": false,
 "bias_reason": "",
 "min_confidence": 0.65,
 "output_key": "product_engagement_depth",
 "redundancy_group_id": "GRP_PAGE_ENGAGEMENT",
 "conflict_penalty_multiplier": 0.5,
 "required_supporting_layers": [],
 "failure_history": [
 {
 "scenario": "Black Friday",
 "failure_mode": "PageValues spike inflates depth_score",
 "lesson": "Attach promo_context_warning when SpecialDay > 0.5"
 }
 ],
 "_llm_instructions": {
 "category_retrieval_logic": "Retrieve when context involves
 product page dwell time, multi-product engagement patterns,
 or high-investment browsing signals.",
 "execution_phase": "Phase Functional — compute first.",
 "redundancy_note": "Member of GRP_PAGE_ENGAGEMENT (cap: 35%).
 Skip if group contribution already at cap.",
 "llm_supplement": "Fully deterministic — no LLM call required.",
 "bias_handling": "No bias risk flag. Standard scoring applies.",
 "constraint_template": "CONSTRAINT:\n[CONDITION] ...\n
 [FAILURE_MODE] ...\n[VERIFICATION_TEST] ..."
 }
}
\end{verbatim}

\section{Complete Element Library Inventory -- behavior\_elements.json}
\label{app:element_inventory}

The following table lists all 24 behavioural elements in the current library. Elements E01–E11 constitute the core library validated in the deterministic engine (\S\ref{subsec:element_ref}). Elements E12–E24 were added during library expansion to improve OSPI signal coverage; these elements participate in the LLM engine's Phase 1 routing but are not used by the deterministic engine's hardcoded dispatch table. All 24 elements are fully structured with computation formula, three-part constraint, FBS mapping, failure history, and \texttt{\_llm\_instructions} fields. None of the 24 elements currently has a second variant (\texttt{variants = 1} for all), representing a known coverage gap for edge-case adaptation.

{\small
\setlength{\tabcolsep}{3pt}
\begin{longtable}{@{} l >{\raggedright\arraybackslash}p{0.34\textwidth} l >{\raggedright\arraybackslash}p{0.36\textwidth} c c l @{}}
\caption{Complete element library inventory (24 elements). $\dagger$ = bias\_risk\_flag=True; (LLM) = llm\_fallback active.} \\
\toprule
\# & Element Name & Layer & Redundancy Group & Bias & min\_conf & Status \\
\midrule
\endfirsthead

\caption[]{Complete element library inventory (continued).} \\
\toprule
\# & Element Name & Layer & Redundancy Group & Bias & min\_conf & Status \\
\midrule
\endhead

\bottomrule
\endlastfoot

E01 & Product\_\allowbreak Engagement\_\allowbreak Depth & Func. & GRP\_\allowbreak PAGE\_\allowbreak ENGAGEMENT & --- & 0.65 & Core \\
E02 & Bounce\_\allowbreak Exit\_\allowbreak Commitment & Func. & GRP\_\allowbreak SESSION\_\allowbreak QUALITY & --- & 0.60 & Core \\
E03 & Page\_\allowbreak Value\_\allowbreak Conversion\_\allowbreak Signal & Func. & GRP\_\allowbreak PAGE\_\allowbreak ENGAGEMENT & --- & 0.70 & Core \\
E04 & Promotional\_\allowbreak Context\_\allowbreak Sensitivity & Func. & GRP\_\allowbreak CONTEXT\_\allowbreak MODIFIERS & $\dagger$ & 0.60 & Core \\
E05 & Research\_\allowbreak Decision\_\allowbreak Style & Inter. & GRP\_\allowbreak USER\_\allowbreak IDENTITY & --- & 0.60 & Core \\
E06 & Traffic\_\allowbreak Source\_\allowbreak Intent\_\allowbreak Score & Inter. & GRP\_\allowbreak USER\_\allowbreak IDENTITY & --- & 0.65 & Core \\
E07 & Visitor\_\allowbreak Loyalty\_\allowbreak Commitment & Inter. & GRP\_\allowbreak USER\_\allowbreak IDENTITY & --- & 0.70 & Core \\
E08 & Search\_\allowbreak Intent\_\allowbreak Semantic\_\allowbreak Inference & Inter. & GRP\_\allowbreak USER\_\allowbreak IDENTITY & (LLM) & 0.45 & Core \\
E09 & Session\_\allowbreak Value\_\allowbreak Composite\_\allowbreak Index & Syst. & GRP\_\allowbreak SESSION\_\allowbreak QUALITY & --- & 0.60 & Core \\
E10 & Device\_\allowbreak Ecosystem\_\allowbreak Pattern & Cont. & GRP\_\allowbreak CONTEXT\_\allowbreak MODIFIERS & $\dagger$(LLM) & 0.55 & Core \\
E11 & Temporal\_\allowbreak Context\_\allowbreak Signal & Cont. & GRP\_\allowbreak CONTEXT\_\allowbreak MODIFIERS & --- & 0.60 & Core \\
E12 & Promo\_\allowbreak Context\_\allowbreak Page\_\allowbreak Value\_\allowbreak Drift & Func. & GRP\_\allowbreak CONTEXT\_\allowbreak MODIFIERS & --- & 0.60 & Exp. \\
E13 & Value\_\allowbreak Signal\_\allowbreak Stability\_\allowbreak Index & Func. & GRP\_\allowbreak SESSION\_\allowbreak QUALITY & --- & 0.65 & Exp. \\
E14 & Exit\_\allowbreak Rate\_\allowbreak Promo\_\allowbreak Context\_\allowbreak Modulated\_\allowbreak Index & Func. & GRP\_\allowbreak CONTEXT\_\allowbreak MODIFIERS & --- & 0.62 & Exp. \\
E15 & Bounce\_\allowbreak Rate\_\allowbreak Traffic\_\allowbreak Intent\_\allowbreak Filter & Func. & GRP\_\allowbreak SESSION\_\allowbreak QUALITY & --- & 0.62 & Exp. \\
E16 & Product\_\allowbreak Page\_\allowbreak Engagement\_\allowbreak Intensity & Func. & GRP\_\allowbreak PAGE\_\allowbreak ENGAGEMENT & --- & 0.65 & Exp. \\
E17 & Device\_\allowbreak Ecosystem\_\allowbreak Complexity\_\allowbreak Index & Cont. & GRP\_\allowbreak USER\_\allowbreak IDENTITY & --- & 0.62 & Exp. \\
E18 & Promo\_\allowbreak Context\_\allowbreak Search\_\allowbreak Drift & Inter. & GRP\_\allowbreak CONTEXT\_\allowbreak MODIFIERS & --- & 0.60 & Exp. \\
E19 & Promo\_\allowbreak Contextual\_\allowbreak Loyalty\_\allowbreak Index & Syst. & GRP\_\allowbreak USER\_\allowbreak IDENTITY & --- & 0.60 & Exp. \\
E20 & Promo\_\allowbreak Duration\_\allowbreak Compression\_\allowbreak Index & Func. & GRP\_\allowbreak SESSION\_\allowbreak QUALITY & --- & 0.62 & Exp. \\
E21 & Research\_\allowbreak Depth\_\allowbreak From\_\allowbreak Page\_\allowbreak Diversity & Inter. & GRP\_\allowbreak SESSION\_\allowbreak QUALITY & --- & 0.62 & Exp. \\
E22 & Decision\_\allowbreak Delay\_\allowbreak From\_\allowbreak Engagement\_\allowbreak Timing & Inter. & GRP\_\allowbreak SESSION\_\allowbreak QUALITY & --- & 0.60 & Exp. \\
E23 & Cross\_\allowbreak Device\_\allowbreak Research\_\allowbreak Span & Inter. & GRP\_\allowbreak USER\_\allowbreak IDENTITY & --- & 0.58 & Exp. \\
E24 & Cross\_\allowbreak Device\_\allowbreak Traffic\_\allowbreak Intent\_\allowbreak Prior & Inter. & GRP\_\allowbreak USER\_\allowbreak IDENTITY & --- & 0.60 & Exp. \\
\end{longtable}}

\subsection{Redundancy Group Membership and Cap Configuration}
\label{subapp:redundancy_groups}

The four redundancy groups define the anti-inflation architecture. Group caps are read from the library metadata field \texttt{valid\_redundancy\_groups} at runtime. The core 11-element execution order (E01$\rightarrow$E11) is also read from library metadata, ensuring that E09 (Systemic layer, depends on five upstream outputs) is always computed last among the core elements.

\begin{table}[h]
\centering
\caption{Redundancy group configuration.}
\label{tab:redundancy_groups}
\small
\begin{tabularx}{\textwidth}{@{}l l X X X@{}}
\toprule
Group & Cap ($C_g$) & Core Members  & Expanded Members  & Risk \\
\midrule
GRP\_PAGE\_ENGAGEMENT & 0.35 & E01, E03 & E16 & Low: 3 members, generous cap \\
GRP\_SESSION\_QUALITY & 0.30 & E02, E09 & E13, E15, E20, E21, E22 & High: 7 members; cap saturates quickly \\
GRP\_USER\_IDENTITY & 0.30 & E05--E08 & E17, E19, E23, E24 & Critical: 8 members; E07 dominates \\
GRP\_CONTEXT\_MODIFIERS & 0.20 & E04, E10, E11 & E12, E14, E18 & High: 6 members; tightest cap \\
\bottomrule
\end{tabularx}
\end{table}

The group imbalance — particularly GRP\_USER\_IDENTITY with 8 members against a 30\% cap — is the primary structural factor behind the repeated selection of engagement-only elements (E01, E02, E05) as key evidence in false positive sessions. In approximately 85\% of OSPI sessions (Returning Visitors), E07 outputs loyalty\_score $\in [0.70, 0.88]$, consuming a disproportionate portion of the group quota and suppressing E05, E06, and E08 contributions. Future work should either raise the GRP\_USER\_IDENTITY cap or reduce the number of members to restore balance.

\section{Prompt Architecture for LLM-Driven Inference}
\label{app:prompt_arch}

The LLM-driven inference engine uses three distinct system prompt files, each stored as an external markdown document in the \texttt{prompts/system/} directory. This separation allows prompt modification without code changes; updated prompts take effect on the next process start. A fourth file (\texttt{prompts/global\_constraints.md}) is automatically prepended to every LLM call as a cross-cutting constraint document.

\subsection{Prompt File Summary}
\label{subapp:prompt_files}

\begin{table}[H]
\centering
\caption{Prompt file summary for LLM engine.}
\label{tab:prompts}
\small
\begin{tabularx}{\textwidth}{@{}l l X l@{}}
\toprule
File & Phase Used & Key Instructions & Token Budget \\
\midrule
element\_selection.md & Phase 1 & Select 3-9 relevant elements; apply routing\_hint; enforce order constraints & $\sim$2k--3k \\
element\_deep\_analysis.md & Phase 2 & Execute formula step-by-step; check constraints; evaluate failure\_history & $\sim$4k--8k \\
masked\_field\_inference.md & Exp-A & Enforce field masking; three-step inference protocol; output structured JSON & $\sim$1.5k--2k \\
global\_constraints.md & All calls & Output format enforcement; prohibited patterns; confidence standards & $\sim$500 \\
\bottomrule
\end{tabularx}
\end{table}

\subsection{On-Demand Element Loading Architecture}
\label{subapp:on_demand_loading}

This Phase 1 $\rightarrow$ Phase 2 selective loading architecture reflects principles introduced in SPARK\_AI \citep{maric2026spark}, a prompt-orchestrated system that governs LLM reasoning through stateful, process-oriented workflows rather than ad hoc generation. SPARK\_AI demonstrates that structured prompt orchestration significantly improves LLM reasoning reliability and interpretability. SemantiClean's element library serves an analogous role to SPARK\_AI's structured state: grounding LLM inference in pre-validated behavioural knowledge rather than free-form text generation \citep{maric2026spark}.

A key design decision in the engine is the separation of Phase 1 (lightweight routing) and Phase 2 (full element payload). Rather than loading all 24 element definitions into every LLM call — which would consume approximately 16,000+ tokens for a naive full-library dump — the engine implements on-demand selective loading:

\begin{table}[h]
\centering
\caption{On-demand element loading token budget.}
\label{tab:loading_tokens}
\small
\begin{tabularx}{\textwidth}{@{}l X X l@{}}
\toprule
Step & Python Action & LLM Receives & Token Cost \\
\midrule
Phase 1 & Load \_summary + index & Names, descriptions, tags, hints ($\sim$6 KB) & $\sim$2k--3k \\
Phase 2 & Fetch full JSON for selected only & Complete formula, constraints, variants, failure history & $\sim$4k--8k \\
Phase 3 & Python aggregation & (none) & 0 tokens \\
\bottomrule
\end{tabularx}
\end{table}

This architecture reduces per-session token consumption from $\sim$16,000+ tokens (full library dump) to approximately 6,000--13,000 tokens depending on how many elements Phase 1 selects. The critical property is that no element definition is disclosed to the LLM before Phase 1 routing confirms its relevance, preserving the semantic integrity of the element selection decision and preventing the LLM from being influenced by element content it was not intended to use.

\subsection{Masked Field Inference Protocol (Exp-A)}
\label{subapp:masked_protocol}

The \texttt{masked\_field\_inference.md} prompt implements a three-step protocol specifically designed for the experimental validation described in \S\ref{subsec:exp_a}:

\paragraph{Step 1 — Signal Extraction.} The LLM computes each element's signal value by applying \texttt{computation.formula} to the visible session fields. If a required field is absent due to masking, \texttt{signal\_value} is set to 0.0 and \texttt{status} is set to 'skipped'. If partial fields allow estimation, \texttt{status} is set to 'degraded' and confidence is capped.

\paragraph{Step 2 — Multi-Element Aggregation.} Signal values are weighted by element confidence and accumulated toward label scores, subject to redundancy group caps. Conflicting signals from bias-flagged elements trigger TieredPenaltyCalculator application.

\paragraph{Step 3 — Decision Output.} The prompt enforces JSON-only output containing \texttt{element\_signals} (per-element signal value, confidence, derivation, status), \texttt{prediction} (target label or 'uncertain'), \texttt{confidence} (0--1), \texttt{decision\_reason} (citing specific element names and values), and \texttt{key\_evidence} (top-3 contributing elements).

The prompt explicitly prohibits use of any field listed in the \texttt{\_\_masked\_fields\_\_} metadata tag. This prohibition is enforced at the prompt level; the session dictionary itself also has the \texttt{Revenue} key removed by Python before the LLM call, providing a defense-in-depth approach to field masking.

\end{document}